\documentclass[conference]{IEEEtran}
\IEEEoverridecommandlockouts
\usepackage{cite}
\usepackage{amsmath,amssymb,amsfonts}
\usepackage{algorithmic}
\usepackage{graphicx}
\usepackage{textcomp}
\usepackage{tablefootnote}
\usepackage{xcolor}
\usepackage{float}
\usepackage[colorlinks=true, allcolors=blue]{hyperref}
\def\BibTeX{{\rm B\kern-.05em{\sc i\kern-.025em b}\kern-.08em
    T\kern-.1667em\lower.7ex\hbox{E}\kern-.125emX}}
\begin{document}
\thispagestyle{empty}

{\LARGE \textbf{IEEE Copyright Notice}}

\vspace{1cm} 

{\small
\noindent \textcopyright\ 2024 IEEE. Personal use of this material is permitted. Permission from IEEE must be obtained for all other uses, in any current or future media, including reprinting/republishing this material for advertising or promotional purposes, creating new collective works, for resale or redistribution to servers or lists, or reuse of any copyrighted component of this work in other works.
}

\vspace{1cm} 

{\small
\noindent Accepted to be Published in: Proceedings of the 2025 IEEE International Conference on Image Processing Applications and Systems (IPAS 2025), 9-11 January, 2025, Lyon, France.
}

\title{Segment Anything for Dendrites from Electron Microscopy\\
\thanks{* Corresponding Authors}
}
\author{
\begin{minipage}{0.27\textwidth}
    \centering
    1\textsuperscript{st} Zewen Zhuo *\\
    \textit{A.I. Virtanen Institute} \\
    \textit{for Molecular Sciences}\\
    \textit{University of Eastern Finland}\\
    Kuopio, Finland\\zewen.zhuo@uef.fi
\end{minipage}

\begin{minipage}{0.23\textwidth}
    \centering
    2\textsuperscript{nd} Ilya Belevich \\
    \textit{Electron Microscopy Unit} \\
\textit{Institute of Biotechnology}
\\\textit{University of Helsinki}
\\Helsinki, Finland\\ilya.belevich@helsinki.fi
\end{minipage}
\begin{minipage}{0.27\textwidth}
    \centering
    3\textsuperscript{rd} Ville Leinonen \\\textit{Department of Medicine} \\
\textit{Faculty of Health Sciences}
\\\textit{University of Eastern Finland}
\\Kuopio, Finland\\ville.leinonen@uef.fi
\end{minipage}
\begin{minipage}{0.23\textwidth}
    \centering
    4\textsuperscript{th} Eija Jokitalo \\
    \textit{Electron Microscopy Unit} \\
\textit{Institute of Biotechnology}
\\\textit{University of Helsinki}
\\
Helsinki, Finland\\eija.jokitalo@helsinki.fi
\end{minipage}

\\
\\

\begin{minipage}{0.3\textwidth}
    \centering
    5\textsuperscript{th} Tarja Malm \\
    \textit{A.I. Virtanen Institute} \\
\textit{for Molecular Sciences}
\\\textit{University of Eastern Finland}
\\
Kuopio, Finland\\tarja.malm@uef.fi
\end{minipage}
\begin{minipage}{0.3\textwidth}
    \centering
    6\textsuperscript{th} Alejandra Sierra \\
    \textit{A.I. Virtanen Institute} \\
\textit{for Molecular Sciences}
\\\textit{University of Eastern Finland}
\\
Kuopio, Finland\\alejandra.sierralopez@uef.fi 
\end{minipage}
\begin{minipage}{0.3\textwidth}
    \centering
    7\textsuperscript{th} Jussi Tohka * \\
    \textit{A.I. Virtanen Institute} \\
\textit{for Molecular Sciences}
\\\textit{University of Eastern Finland}
\\
Kuopio, Finland\\jussi.tohka@uef.fi
\end{minipage}
}

\maketitle

\begin{abstract}
Segmentation of cellular structures in electron microscopy (EM) images is fundamental to analyzing the morphology of neurons and glial cells in the healthy and diseased brain tissue. Current neuronal segmentation applications are based on convolutional neural networks (CNNs) and do not effectively capture global relationships within images. Here, we present DendriteSAM, a vision foundation model based on Segment Anything, for interactive and automatic segmentation of dendrites in EM images. The model is trained on high-resolution EM data from healthy rat hippocampus and is tested on diseased rat and human data. Our evaluation results demonstrate better mask quality compared to the original and other fine-tuned models, leveraging the features learned during training. This study introduces the first implementation of vision foundation models in dendrite segmentation, paving the path for computer-assisted diagnosis of neuronal anomalies.
\end{abstract}


\section{Introduction}
\label{sec:intro}
Advancements in serial block-face scanning electron microscopy (SBF-SEM) allow researchers to capture brain tissue images at nanometer resolution across hundreds of micrometers \cite{zheng2018complete,hildebrand2017whole}. Segmenting complex ultrastructures such as axons, soma, and dendrites from electron microscopy (EM) data is critical for analyzing morphological parameters or reconstructing 3-dimensional (3D) neuronal networks. Dendrites, in particular, are intricate and varied, with numerous studies linking dendritic pathology to neurodegenerative diseases such as Alzheimer's and Parkinson's \cite{baloyannis2009dendritic,penzes2011dendritic,villalba2010striatal,kweon2017cellular}. While manual dendrite annotation has long been the benchmark, it is labor-intensive and time-consuming \cite{berning2015segem,abdollahzadeh2019automated}. Deep learning, especially convolutional neural networks (CNNs), has improved segmentation efficiency, but CNNs struggle with generalization across diverse datasets. Inspired by the success of large language models (LLMs), there is growing interest in applying foundation models to computer vision tasks.

The name, foundation models, originates from their fundamentally
central yet inherently incomplete nature, serving as a robust foundation that can be adapted to downstream tasks \cite{bommasani2021opportunities}. Models such as GPT-4 \cite{achiam2023gpt}, PaLM-2 \cite{anil2023palm}, and LLaMA \cite{touvron2023llama} have benefited from the sheer scale of data they have been trained on and have demonstrated to generalize beyond the training data. Research on foundation models in computer vision, such as the segment anything model (SAM)\cite{kirillov2023segment} and segment-everything-everywhere model (SEEM) \cite{zou2024segment}, demonstrates exceptional versatility in segmentation tasks. Notably, trained on around 1 billion masks from 11 million images, SAM is a state-of-the-art (SOTA) segmentation model. 

SAM has been widely studied and integrated into various downstream tasks, such as zero-shot abdominal organ segmentation in computed tomography (CT) \cite{roy2023sam} and skull-stripping in brain magnetic resonance imaging (MRI) \cite{mohapatra2023sam}. SAM's robustness  has been found to be remarkable against various types of corruption, with the exception of blur-related corruption\cite{huang2023robustness}. 
In spite of its generalizability gained from large training data, SAM's performance degrades drastically when handling objects in medical imaging with blurred boundaries or weak contrast \cite{zhou2023can}. Therefore, further fine-tuning is essential for the precise segmentation of such objects. Here, we introduce DendriteSAM, a vision foundation model for segmenting dendrites from large-scale EM volumes, and evaluate it internally as well as externally and across subjects and species. Our contributions are:
\begin{itemize}
    \item we present the first application of a vision foundation model specialized in dendrite segmentation; 
    \item we evaluate DendriteSAM's robustness both internally and externally through comprehensive quantitative and qualitative evaluation and demonstrate improved  segmentation performance compared to SOTA segmentation models;
    \item we investigate the impact of different image pre-processing methods on dendrite segmentation.
\end{itemize}
\section{RELATED WORK}

Proposed in 2023, SAM is a groundbreaking image segmentation model. It offers SOTA performance by exploiting a gigantic scale of training data, enabling it to handle a wide range of segmentation tasks with high accuracy and efficiency. The model incorporates three 
fundamental modules: an image encoder, a prompt encoder, and a mask decoder.

The image encoder relies on vision transformers (ViTs) \cite{dosovitskiy2020image} and masked autoencoders (MAE) \cite{he2022masked} to translate input images into multi-dimensional embeddings. SAM’s image encoder varies in the number of parameters and is available in three different scales: ViT-Base (ViT-B), ViT-Large (ViT-L), and ViT-Huge (ViT-H). The prompt encoder supports sparse prompts (points, boxes, text) and dense prompts (masks) as input, indicating what to segment in the images. After computing the mask embeddings and prompt embeddings, a lightweight mask decoder is designed to efficiently map the image embeddings, prompt embeddings, and output tokens to masks with confidence scores. The model was evaluated in multiple downstream tasks, such as edge detection, instance segmentation, and object proposal generation through zero-shot learning\cite{kirillov2023segment}.

SAM's utilization on medical images, including zero-shot and few-shot learning, has been broadly investigated. A seminal work in the medical context is MedSAM \cite{ma2024segment}, a vision foundation model designed for universal medical image segmentation. MedSAM was trained on 1,570,263 medical image-mask pairs derived from 10 medical imaging modalities, including MRI, CT, X-ray, and ultrasound, among others. This extensive training endowed the model with the ability to segment objects in various types of medical images and anomalies. In addition, it compensated for SAM's limitation in segmenting medical objects, such as polyps \cite{zhou2023can}. The model incorporated the architecture from SAM and initialized the parameters with pretrained SAM using the ViT-B image encoder. To simplify training, only box prompts were considered, and the prompt encoder was frozen in the training phase. This model was evaluated using dice similarity coefficient (DSC) and demonstrated narrower DSC contribution compared to default SAM as well as modality-wise CNN models, revealing MedSAM's robustness across a variety of tasks. On the other hand, the fixed mask encoder and solitary box prompt hindered the model from iteratively learning and improving from input prompts, compromising the model's performance to a degree.

In addition to universal medical image segmentation, segmentation specialized in microscopy images is a crucial component in histological and anatomical analysis. Micro\_SAM is seminal in applying a vision foundation model to light microscopy (LM) and EM images \cite{archit2023segment}. Micro\_SAM integrated mask, box, and point prompts during training and proposed an iterative training scheme to fine-tune SAM on microscopy data. Additionally, it systematically trained generalist models for LM and EM. The evaluation results showed impressive improvements in LM relative to the original SAM. Lastly, an annotation tool based on Napari \cite{ahlers_2023_8115575} was developed for model-assisted annotation, providing multiple options for interactive annotation on microscopy data. Nevertheless, the study faced challenges with the hierarchical structures in EM images and failed to develop a model that generalizes well to all EM data. Furthermore, the model trained on EM was not evaluated for automatic instance segmentation, limiting the assessment of the model's performance in fully automated workflows.

\section{METHOD}
\label{sec:methods}
\subsection{Datasets}

Aiming to develop a vision foundation model in segmenting dendrites, we curated EM data acquired using SBF-SEM. This technique allows for high-resolution 3D imaging by sequentially imaging the block face and sectioning ultra-thin slices of the tissue specimen \cite{smith2019serial}. SBF-SEM has been used for reconstructing detailed 3D structures of biological tissues and
complex materials \cite{PHELPS2021759}. Here, we acquired three 
datasets (A, B, and C) with a cutting interval of 40 nm, summarized in Table \ref{tab:Data_overview}. Dataset A was from the CA1 region of the hippocampus of a healthy rat, serving as a control
to understand normal dendrite structures. Dataset B was obtained from the same region as dataset A, from a rat after status epilepticus induced by pilocarpine. Dataset C was obtained from a biopsy of cortical layer II of a human patient diagnosed with idiopathic normal pressure hydrocephalus during shunt surgery. Animal procedures were approved by Committee of the Provincial Government of Southern Finland, 
following the European Community Council Directives 86/609/EEC. 
The collection of human data was approved by the Research Ethics Committee of the Northern Savo Hospital District and participant gave written informed consent.

\begin{table}[h]
\centering
\caption{Overview of Data Utilized} 
\label{tab:Data_overview}  
\resizebox{\columnwidth}{!}{
\begin{tabular}{|c|c|c|c|c|} 
\hline
Dataset&Pixel Size&Width $\times$ Height & Slices & Employed  \\
\hline
 A & 15 $\times$ 15 \(nm^2\) & 3100 $\times$ 3100 & 1044 &Training \& Evaluation\\
\hline
 B  &15 $\times$ 15 \(nm^2\)& 3100 $\times$ 3100 & 698 &Evaluation  \\
\hline
C &10 $\times$ 10 \(nm^2\) & 4096 $\times$ 4096 & 697 &Evaluation  \\
\hline
\end{tabular}
}
\end{table}

\begin{figure}[H]
    \centering
    \includegraphics[width=0.48\textwidth]{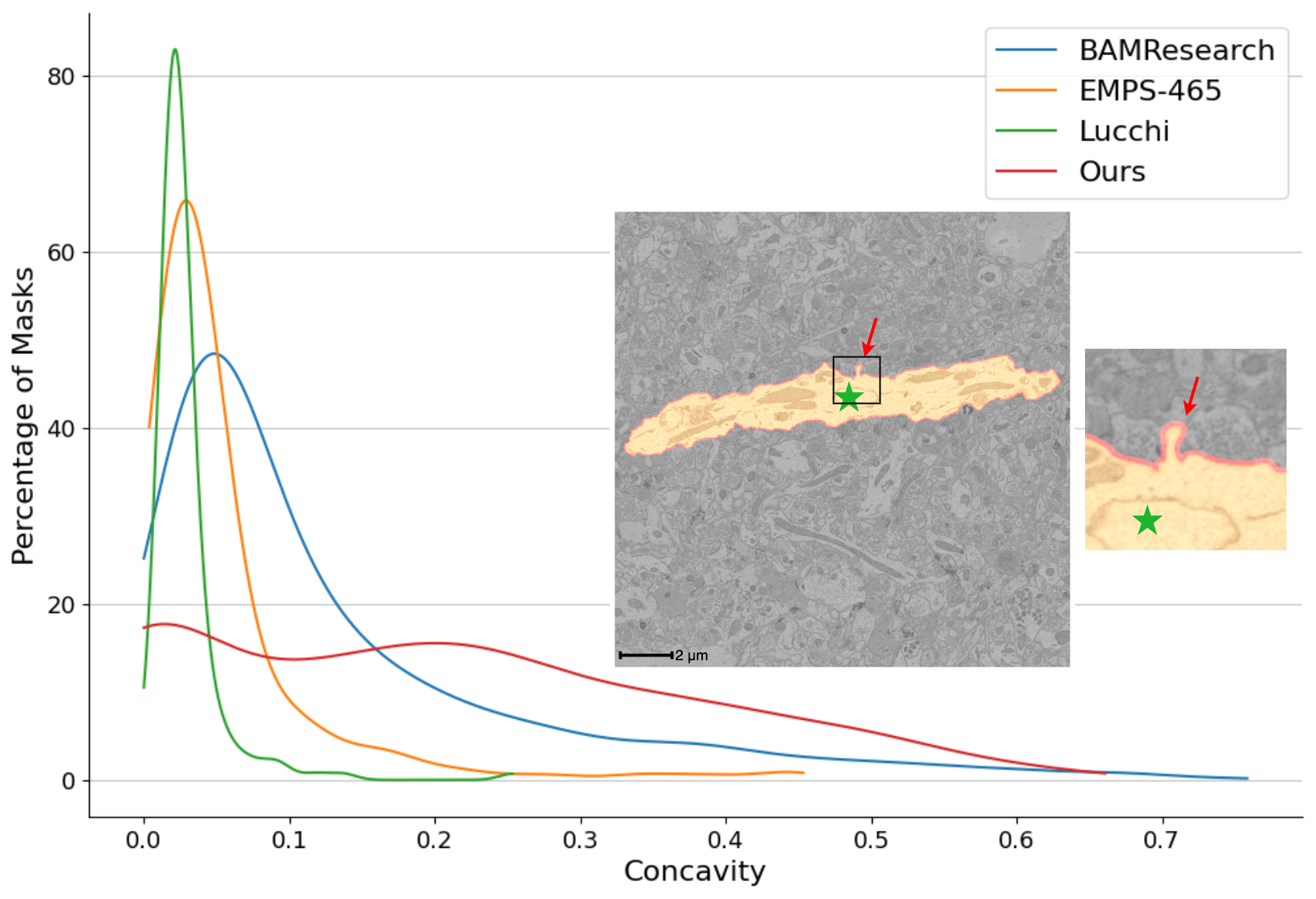}
    \caption[Concavity]{Concavity Distribution: The x-axis and y-axis represent the value of concavity and the corresponding percentage of masks in sampled images, respectively. The green star and red arrow in the subplot indicate the dendrite and spine structure, respectively.}
    \label{fig:concavity}
\end{figure}

For training, we sparsely annotated 100 slices of images from dataset A to balance the need for high-quality annotations with time spent for annotation. Both dendrites and their associated dendritic spines were included and referred to as dendrites in this study. The green star in Figure \ref{fig:concavity} subplot indicates the dendrite and the red arrow points to the dendritic spine. The annotation was performed by Z.Z. under the supervision of A.S., an expert in neuroanatomy. Following annotation, the masks were validated by A.S. before being referred to as ground truth (GT) masks.
Before model development, we investigated the object complexity by calculating the concavity
\begin{equation}
    {concavity} = 1- \frac{mask\, area}{mask\, convex\, hull\, area}
    \label{eq:concavity}
\end{equation}
and compared it to public EM datasets (BAMResearch \cite{ruhle2021workflow}, EMPS-465 \cite{yildirim2021bayesian}, and Lucchi \cite{lucchi2012structured})
by randomly sampling 10 images from each dataset. The results in Figure \ref{fig:concavity} suggest that our dataset has a high proportion of objects with concavity values exceeding 0.2, indicating that the objects in our data are more complex than those in public datasets. For evaluation, we annotated another five randomly sampled images for each dataset in Table \ref{tab:Data_overview}. Among five images, two were used for hyperparameter search in automatic inference, and three were used for both interactive and automatic inference.

\subsection{Model Architecture and Training Protocols}
\begin{figure}[H]
    \centering
    \includegraphics[width=0.48\textwidth]{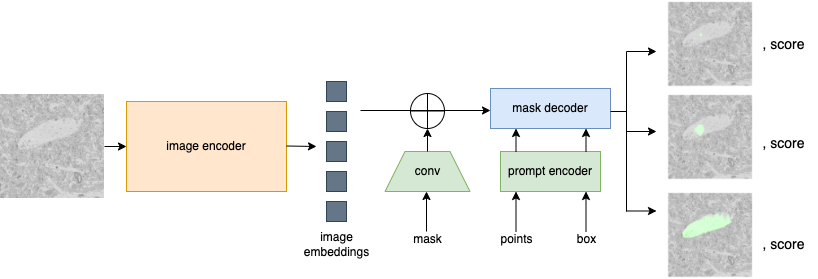}
    \caption[Model Architecture]{Model Architecture}
    \label{fig:model}
\end{figure}

We employed the model architecture from SAM and incorporated mask, point, and box prompts. Text prompts were excluded due to their unavailability in the released model. The image encoder in our study was based on ViT-B and ViT-L, since researchers have proved that ViT-H has only marginal improvements over ViT-L while being computationally expensive \cite{kirillov2023segment}. The model was designed to handle ambiguity when using a single point prompt, with an estimated intersection over union (IoU) score to rank the masks, as in Figure \ref{fig:model}.

The iterative training scheme was adopted from \cite{archit2023segment}, which was activated with a foreground point prompt or a bounding box prompt. Subsequently, it iteratively incorporated predicted masks as well as additional sampled point prompts during training. Due to SAM's nature of predicting both masks and confidence scores, the dice loss function and the L2 loss function were employed to compute the losses. Specifically, the dice loss was applied to calculate the loss between mask predictions and GT masks, and the L2 loss was used to calculate the loss between estimated IoU and true IoU. Afterwards, all losses were averaged across all iterative steps. The model was optimized using an Adam optimizer \cite{kingma2014adam} with an initial learning rate of 10\textsuperscript{-5} ($\beta_1$ = 0.9, $\beta_2$ = 0.999) and was trained on an A100 GPU with 40 GB of VRAM for 100 epochs without early stopping. 

\subsection{Evaluation Metrics}
We followed the evaluation paradigm for microscopy image segmentation from \cite{caicedo2019nucleus},  which is based on quantifying the object-level error. The metric for indicating the inference mask quality is defined by:
\begin{equation}
    {Quality} = \frac{1}{|T|} \sum_{t \in T} \frac{TP(t)}{TP(t)+FP(t)+FN(t)}
    \label{equ:metrics1}
\end{equation}
where \textit{TP}, \textit{FP}, and \textit{FN} stand for true positives, false positives, and false negatives derived from object matching by computing IoU pairs and differentiating correct/error segmentation with an IoU threshold $t \in T = \{0.5,0.55,\ldots,1\}$. The quality value is between 0 and 1, with a higher value indicating better mask quality. In \cite{caicedo2019nucleus}, they utilized this metric to evaluate nuclei segmentation, where average quality scores above 0.6 for simpler nucleus structures under a lower threshold range were considered outstanding.

Furthermore, we evaluated the annotated mask similarity under two annotation modes via DSC and the 95th percentile Hausdorff distance (HD95) in our user study.

\section{RESULTS}
\subsection{Interactive Inference}
For interactive inference, we simulated user input prompts by sampling points or boxes derived from the GT masks.  These prompts were cached in files to ensure reproducibility and facilitate precise comparison across different models. In subsequent graphs, foreground points, background points, and box prompts are denoted as "p," "n," and "bbox," respectively. For example, bbox\_p4\_n8 refers to the combination of a bounding box, 4 foreground points, and 8 background points.
\subsubsection{Quantitative Evaluation}

We first evaluated the impact of image tiling on the model performance. In [\citenum{ma2024segment}], a sliding window was implemented to partition whole-slide pathology images to meet the input size requirement (1024 $\times$ 1024) of SAM, and a similar technique was also employed in another microscopy segmentation study\cite{Bhattiprolu2023}. However, the impact of tiling the image versus directly resizing it was uncertain and needed investigation, especially when dealing with large image sizes.

\begin{figure}[H]
    \centering
    \includegraphics[width=0.45\textwidth]{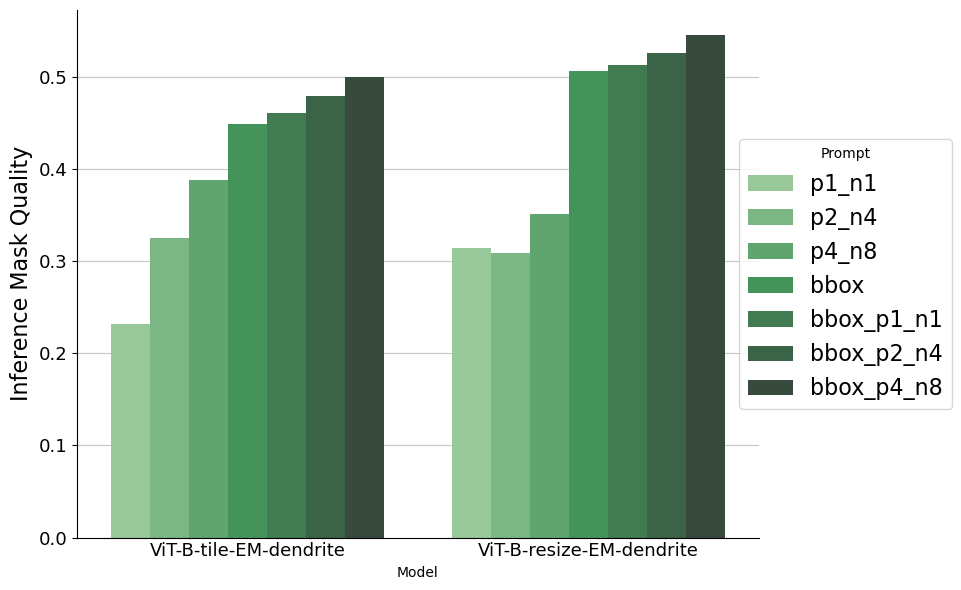}
    \caption[quantitative]{Impact of Image Tiling vs Resizing: On the left, segmentation quality with the model trained with the tiling method  (cropping with a sliding window size 1024 and step size 512), and on the right  with the model trained with the resizing method (resizing the whole image to 1024 $\times$ 1024). In the legend, the digits after "p" and "n" stand for the number of foreground points and background points used, respectively.}
    \label{fig:quantitative1}
\end{figure}

\begin{figure}[H]
    \centering
    \includegraphics[width=0.5\textwidth]{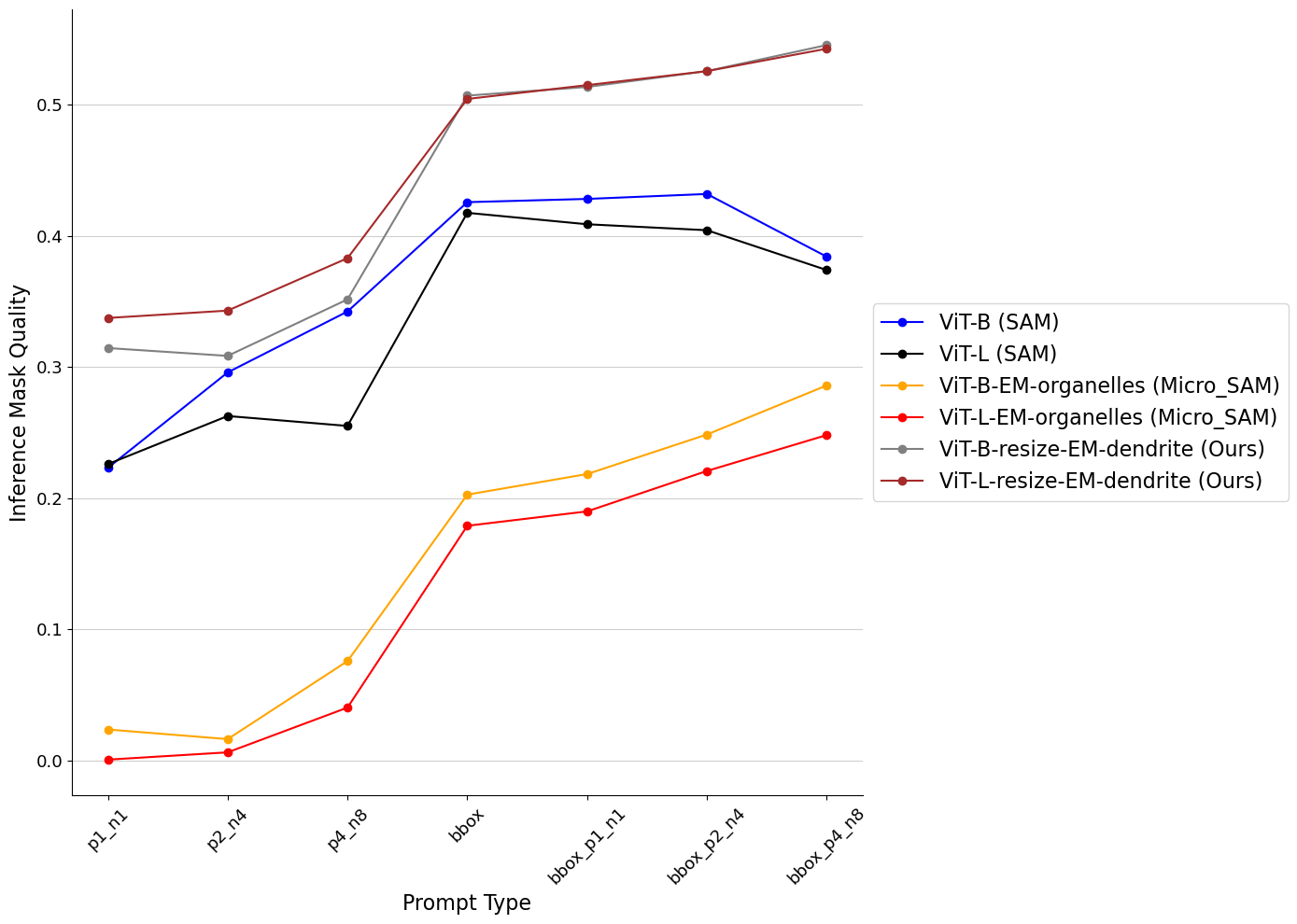}
    \caption[quantitative]{Quantitative Evaluation: The blue and black legends present ViT-B and ViT-L models from SAM. ViT-B-EM-organelles \cite{weight1} and ViT-L-EM-organelles\cite{weight2} stand for the models from Micro\_SAM trained on EM organelles images, and ViT-B-resize-EM-dendrite and ViT-L-resize-EM-dendrite are specialist models that we trained.}
    \label{fig:quantitative2}
\end{figure}

Therefore, we trained two ViT-B models initialized with weights from Micro\_SAM-ViT-B-EM-organelles \cite{weight1} using different methods. Results in Figure \ref{fig:quantitative1} reveal that cropping images generally undermined the model's predictions across all evaluation datasets. For both models, it is evident that using different number of point prompts led to varying degrees of fluctuations in mask quality. Bounding box prompts enhanced predictions in both models compared to point prompts, with approximately 15.8\% and 44.2\% advancements in mask quality compared to the results from the p4\_n8 prompt combination, respectively. The bbox\_p4\_n8 prompt combination yielded the best results at 0.500 and 0.545 for ViT-B-tile-EM-dendrite
and ViT-B-resize-EM-dendrite, respectively.

 Following the comparative study, we decided to conduct training on resized data. One common weakness shared by SAM and MedSAM was their inability to accurately segment thin and elongated branching objects. To address this, we investigated improvements in mask quality across our models, all initialized with Micro\_SAM pretrained weights for EM organelles \cite{weight1,weight2}, relative to both the original SAM and Micro\_SAM. The quantitative
results are illustrated in Figure \ref{fig:quantitative2}. Both fine-tuned models outperformed their counterparts markedly. The most notable enhancement in mask quality was observed when transitioning from point prompts to bounding box prompts. Using the mask quality of bounding box prompts as a benchmark, our fine-tuned ViT-B and ViT-L models exhibited approximately 19.1\% and 20.8\% improvements over the original SAM, and 150.1\% and 181.9\% improvements over Micro\_SAM models, respectively. This is because two Micro\_SAM models used in this study were trained for organelles. Figure \ref{fig:quantitative2} also demonstrates that bounding box prompts and their combinations, which include bounding box prompts with point prompts, outperformed pure point prompts. Considering all prompt types, ViT-L-resize-EM-dendrite slightly outperformed ViT-B-resize-EM-dendrite, and therefore, its predictions were used in the qualitative evaluation.

\begin{table}[h]
\centering
\caption{Qualitative Grading Criteria} 
\label{tab:Grading}    
\begin{tabular}{|l|l|} 
\hline
Scale & Quality \\
\hline
1 & Poor Quality \\
\hline
2 &  Major Part of the Object Missing \\
\hline
3 & Object Moderately Segmented\\
\hline
 4 & Major Part of the Object Segmented  \\
\hline 
5 & Acceptable Quality  \\
\hline 
\end{tabular}

\end{table}
\begin{figure}[H]
    \centering
    \includegraphics[width=0.44\textwidth]{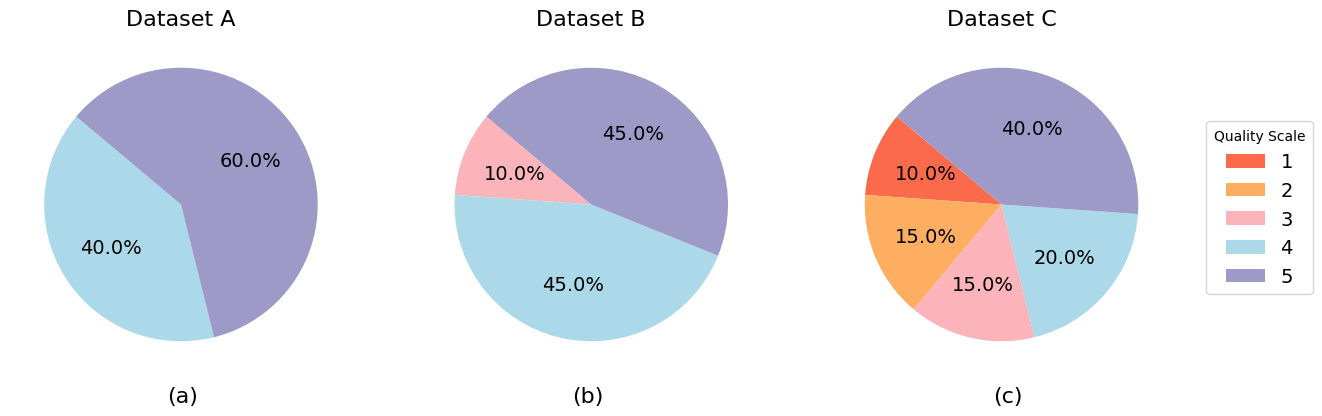}
    \caption[qualitative]{Qualitative Analysis: Expert graded quality of randomly sampled masks predicted by ViT-L-resize-EM-dendrite with bbox\_p4\_n8 prompts.}
    \label{fig:qualitative}
\end{figure}

\subsubsection{Qualitative Evaluation}

 Predicted object masks were sampled from predictions of ViT-L-resize-EM-dendrite with bbox\_p4\_n8 prompts, and their qualities were assessed by A.S.,
 following the criteria outlined in Table \ref{tab:Grading}. The quality score distributions across different evaluation datasets are depicted in Figure \ref{fig:qualitative}. Dataset A, which was an internal evaluation dataset but not used in training, exhibited the highest quality. Predictions on the animal model data, dataset B, generally performed well but encountered challenges with certain objects. In human data, approximately 60\% of sampled objects achieved a quality score of 4 or higher. Although the quality varied in human data, its mask quality degradation is understandable considering the transition from animal data to human data, the influence of the disease, the type of cells imaged, as well as the distinct imaged brain region.

\subsubsection{User Study}

\begin{figure}[H]
    \centering
    \includegraphics[width=0.5\textwidth]{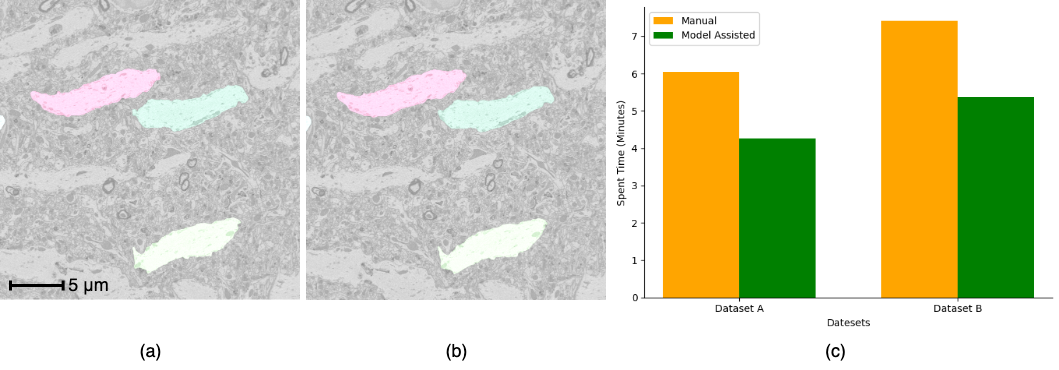}
    \caption[user1]{Quality Comparison in User Study and Efficiency Improvements: (a) masks from fully manual annotation, (b) masks from model-assisted annotation, (c) average time spent per object under different annotation modes.}
    \label{fig:user1}
\end{figure}

\begin{figure}[H]
    \centering
    \includegraphics[width=0.45\textwidth]{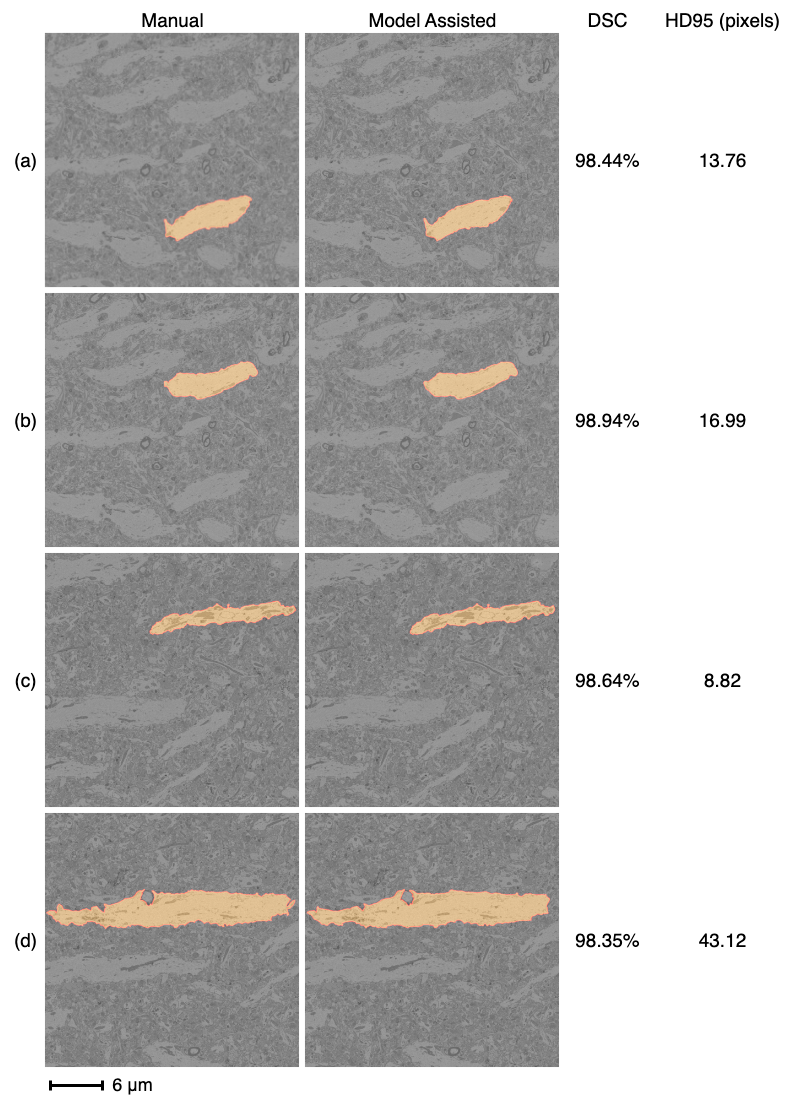}
    \caption[user2]{Mask Similarities in Different Annotation Modes: (a) \& (b) are examples from dataset A, (c) \& (d) are examples from dataset B.}
    \label{fig:user2}
\end{figure}
Given the promising results from both evaluations, we were also interested in testing the model's contribution in model-assisted segmentation of dendrites. Therefore, we implemented a user study where a candidate was required to either annotate the objects fully manually or to refine the masks proposed by the fine-tuned model (ViT-L-resize-EM-dendrite) until they satisfied the user. The user was not limited to any specific type or number of prompts, and annotation was carried out using the annotation tool from Micro\_SAM. Figure \ref{fig:user1} (c) illustrates the time reductions in the model-assisted annotation mode. Although amending the proposed masks still required time, the efficiency had significantly improved. The quality of the masks produced by manual annotation and model-assisted annotation is illustrated in Figure \ref{fig:user1} (a) and (b), respectively. To compare the masks statistically, Figure \ref{fig:user2} presents the similarities between two different annotation strategies using DSC and HD95. The results demonstrated high mask similarities between the two methods.

\subsection{Automatic Inference}
\begin{figure}[H]
    \centering
    \includegraphics[width=0.44\textwidth]{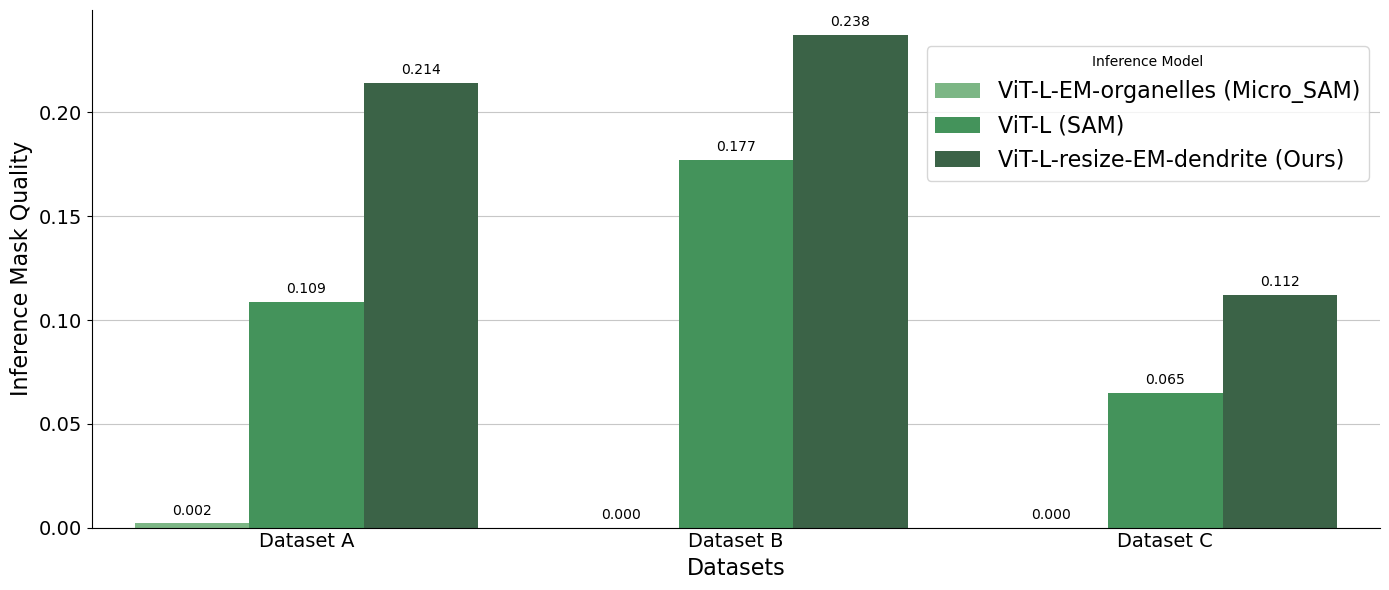}
    \caption[auto]{Quantitative Evaluation of Automatic Inference}
    \label{fig:auto_quantitative}
\end{figure}

\begin{figure}[H]
    \centering
    \includegraphics[width=0.5\textwidth]{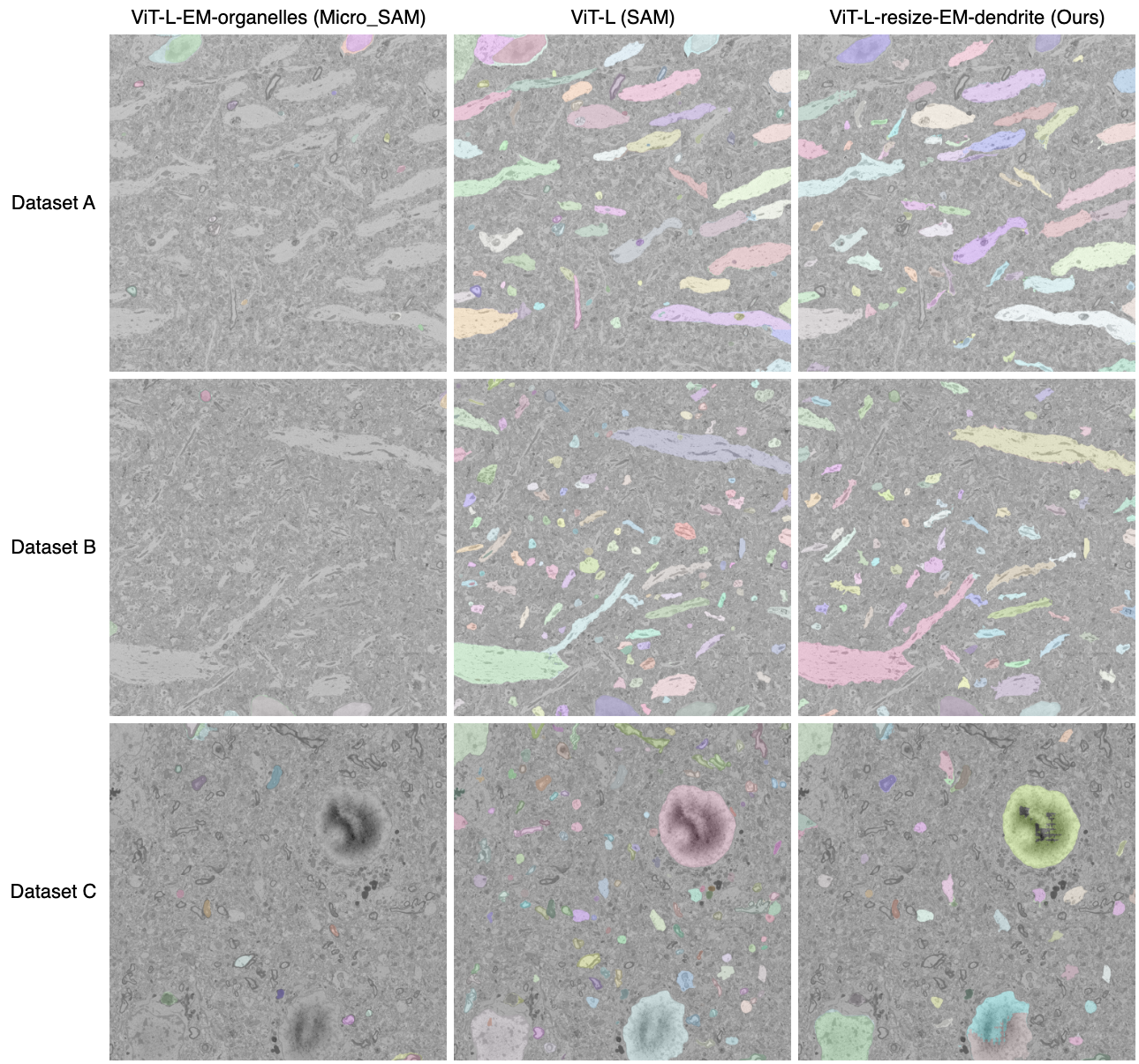}
    \caption[auto2]{Qualitative Analysis for Automatic Inference}
    \label{fig:auto_qualitative}
\end{figure}
In order to complement the absence of automatic inference for EM in Micro\_SAM, we conducted automatic inference on ViT-L-EM-organelles, ViT-L, and ViT-L-resize-EM-dendrite, to compare our model to Micro\_SAM and SAM.  The quantitative results in Figure \ref{fig:auto_quantitative} depict that the model performance was enhanced prominently compared to Micro\_SAM and original SAM research. However, the mask quality was inferior to that of interactive inference where prompts embedded more spatial information of target objects.

To better interpret the automatic instance segmentation result, the mask quality in each model is illustrated in Figure \ref{fig:auto_qualitative}, from which we could see that fewer error predictions were made in our fine-tuned model and instances are more precisely labeled.

\section{CONCLUSION}
The development of the vision foundation model in segmenting dendrites aims to help neuroscientists interactively segment these structures. DendriteSAM marks the first application of a vision foundation model specialized in dendrites and achieved satisfying results in relation to other oft-cited segmentation foundation models in animal data, with around 0.1 and 0.3 mask quality enhancement compared to SAM and Micro\_SAM in interactive inference, respectively. It also presents the impressive capability of vision foundation models in solving complex object segmentation tasks. We also reported the results from the automatic inference and analyzed the impact of image tiling on model performance. Further research could focus on improving the accuracy of automatic segmentation, optimizing the performance in human data through few-shot learning, and quantifying morphological parameters of dendrites. To facilitate reproducibility, the code and DendriteSAM model weights are available at \href{https://github.com/ZE-WEN/DendriteSAM}{https://github.com/ZE-WEN/DendriteSAM}.
\section*{Acknowledgements}

This work was supported in part by the Research Council of Finland (\#323385, \#346934 and \#358944 (Flagship of Advanced Mathematics for Sensing Imaging and Modelling)), and the Jane and Aatos Erkko Foundation. Thanks to Henna Jäntti for help with the user study and the Bioinformatics Center at the University of Eastern Finland for providing computational resources.

\bibliography{report}
\bibliographystyle{ieeetr} 
\end{document}